\pdfoutput=1

\documentclass[11pt]{article}

\usepackage{ACL2023}

\usepackage{times}
\usepackage{latexsym}

\usepackage[T1]{fontenc}

\usepackage[utf8]{inputenc}

\usepackage{microtype}

\usepackage{inconsolata}
\usepackage{graphicx} 
\usepackage{booktabs} 
\usepackage{amsmath}
\usepackage{amssymb}
\usepackage{adjustbox} 

\title{Evaluating and Adapting Large Language Models to Represent Folktales in Low-Resource Languages}

\author{J.A. Meaney, Beatrice Alex \and William Lamb    \\
  School of Languages, Literatures and Cultures, \\
  University of Edinburgh \\
  \texttt{jameaney@ed.ac.uk, b.alex@ed.ac.uk, w.lamb@ed.ac.uk}}

\begin{document}
\maketitle
\begin{abstract}
Folktales are a rich resource of knowledge about the society and culture of a civilisation. Digital folklore research aims to use automated techniques to better understand these folktales, and it relies on abstract representations of the textual data.  Although a number of large language models (LLMs) claim to be able to represent low-resource langauges such as Irish and Gaelic, we present two classification tasks to explore how useful these representations are, and three adaptations to improve the performance of these models. We find that adapting the models to work with longer sequences, and continuing pre-training on the domain of folktales improves classification performance, although these findings are tempered by the impressive performance of a baseline SVM with non-contextual features. 
\end{abstract}

\section{Introduction}
Arguably the biggest development in natural language processing in recent years has been the use of pre-trained large language models (LLMs) such as BERT \citep{devlin2018bert} to transfer the linguistic knowledge from models trained on massive corpora to other tasks, without the need for retraining. These LLMs have unlocked new avenues of investigation for many fields of research, such as digital folklore \citep{lauer2023computational}. Progress has been particularly rapid for high resource languages, such as English, Spanish, Mandarin, as the high threshold of data required by LLMs is easily met by what is available online. However, despite multiple models claiming to represent a selection of low-resource languages, the relative scarcity of data available to train on may have a substantial impact on the model’s ability to output a faithful representation of the language. This in turn can restrict digital folklore research to using older technologies, or to working on corpora from higher-resource languages.

The current project is an ambitious contribution to folklore preservation and analysis which initially focused on digitally rendering thousands of hand-transcribed tales in Irish and Gaelic \citep{sinclair2022handwriting}. With this step complete, we now aim to use cutting-edge language technology to uncover cultural and linguistic links between the folklore traditions of Ireland and Scotland. 

Our dataset presents two challenges: firstly that it comprises two low-resource languages - Irish and Scottish Gaelic. We are interested in measuring how well LLMs represent the languages in our dataset, using the base version of the model with no adaptations. We also experiment with continuing fine-tuning of these base models with a bilingual corpus of folktale - i.e. domain adaptation. The second challenge is that many of our tales exceed the maximum context window allowed by common LLMs. This leads us to the following research questions:

\begin{enumerate}
    \item How well do base models work for low-resource languages? 
    \item Does an increase in maximum sequence length improve performance on this dataset?
    \item Does domain adaptation improve performance? 
\end{enumerate}

 We operationalise these questions by setting up two classification tasks - to predict the type of folktale and the gender of the person who told the folktale. 

\section{Related Work}

Although the field of computational folkloristics is relatively new, predicting the tale type of a folktale has been approached a number of times. 
\citet{nguyen2013folktale} classified the tale type of Dutch folktales using the Learning to Rank algorithm. The features to their system included measures of lexical similarity using TFIDF representations and Jaccard similarity, and subject-verb-object triplets extracted from the text. Their nearest neighbours approach easily outperformed their baselines.

\citet{lindemann1annotation} focused not on ATU prediction, but on predicting one of the elements which determines which ATU number is appropriate - locations within the tale. They annotated a corpus of German folktales for location, with moderate inter-annotator agreement. They then used rule-based and statistical approaches to classify the locations. Their best-performing system was a hybrid Naive Bayes system with some rule-based features. 

\citet{lo2020exploring} experimented on a corpus of 742 West African and Western European folktales, all translated into English. Amongst their experiments, they trained classifiers to distinguish between the African and European tales, using a BoW as input to a multi-layer perceptron, as well as word embeddings as input to a long short-term memory (LSTM) system. The much simpler BoW approach performed significantly better than the more complex LSTM system, and the authors speculate that it was the small size of the dataset that led to this. 

\citet{eklund2023teaching} used SVMs to predict the 10 most populated classes of the English-language ATU-annotated tale collections available online (1,518 texts annotated with 182 ATU tale types). They used TFIDF features as input, and reported F-scores ranging between 0.8 and 1.0 for the ten best populated classes. However they did not report any baseline model, so it is difficult to contextualise their achievements. 	 

Related work on Irish and Scottish Gaelic speech and language processing more widely can be found in \citet{lamb2022goidelic}, including work on handwriting recognition of the folktale data used for this project \cite{sinclair2022handwriting,o-raghallaigh-etal-2022-handwritten}. The current paper is the first study applying computational text classification methods to Irish and Scottish Gaelic folktales.

\section{Dataset}
The Irish and Gaelic folktales dataset is a collection of 4,692 folk tales collected from two national folklore archives. The Irish language data derive from a 40k page subset of the National Folklore Collection of Ireland’s Main Manuscript Collection, hosted at University College Dublin. This collection comprises over 700k manuscript pages of transcribed folklore interviews made by the Irish Folklore Commission in the mid-20th century. \citet{o-raghallaigh-etal-2022-handwritten} describe the digitisation and semi-automatic handwriting recognition (HWR) of our subset of this material, which amounts to 2091 transcription MSS and 3,829,559 words.The average length of the Irish folktales is 1,831.86 words (SD = 1,827.26). 

The Gaelic data come from the University of Edinburgh’s School of Scottish Studies Archives. \citet{sinclair2022handwriting} describe how the dataset was created by scanning and semi-automatically recognising 2601 transcriptions and published versions of Scottish Gaelic folklore, amounting to 3,048,348 words. The average length of the Scottish folktales is 5,869 (SD = 12,843).

This dataset includes extensive metadata, such as the type of tale, and the gender of the interviewee who contributed the tale - these two variables are the focus of the below classification tasks detailed in Section \ref{sec:methodology}. The type of tale  is determined by the ATU number assigned to it \cite{uther2004types}. The ATU scheme is a taxonomy used specifically for International Folktales. Each number represents a distinct tale type and was assigned to our tales by professional folklorists. ATU numbers run from 1-2499 and have three main divisions: 1--299 (Animal Tales), 300--1199 (Ordinary Folktales) and 1199--2499 (Jokes, Anecdotes and Formula Tales).

\subsection{Domain Adaptation Data}\label{domain-data}
There are several aspects of our dataset which likely differ from the data used to train the LLMs we experiment on, this can result in a domain mismatch. As much of our data was collected 30+ years ago, historical language change is a factor, as is the genre of the folktale itself differing from day-to-day speech. For this reason, we experiment with domain adaptation, which continues the pre-training of the LLM in order to help it to learn more about the domain of folktales and style of language used in the past. 

Two additional datasets were used for domain adaptation. One is a 400k-word collection of orthographically-normalised Gaelic folktales derived from the Calum Maclean Collection.\footnote{\url{https://www.calum-maclean-project.celtscot.ed.ac.uk/home/}} These are verbatim transcriptions of fieldwork recordings of Gaelic-speaking tradition bearers in the mid-20th-century. The other, known as `The Schools' Collection',\footnote{\url{https://www.duchas.ie/en/info/cbe}} is a body of folklore taken down by Irish school children from family members and neighbours, also in the mid-20th-century.  

\section{Methodology}\label{sec:methodology}

We set up two classification tasks - gender and ATU prediction. There were over 300 ATU types represented in our dataset, so in order to avoid issues of data sparsity, we binned the tale types in our corpus into these four broad categories: `Animal' (ATU 0-299), `Magic' (ATU 300--745), `Ordinary (non-Magic)' (ATU 746-1,199), and `Jokes' (ATU 1,200-2,499), in addition to one for Tales of Magic (aka `Fairy Tales': ATU 300--749). While all of the Irish tales have been labelled with an ATU number, only 451 Gaelic tales were appropriate to label this way; those without an ATU number were omitted from the classifier.

\begin{table}[h!]
\centering

\begin{tabular}{lll}
\hline
\textbf{Bin}   & \textbf{Gaelic} & \textbf{Irish} \\ \hline
Magic          & 158             & 1149           \\ \hline
Jokes          & 175             & 481            \\ \hline
Ordinary       & 96              & 392            \\ \hline
Animal         & 22              & 69             \\ \hline
\textbf{Total} & 452/2601        & 2091/2091    
\end{tabular}
\caption{Number of tales per ATU bin in Irish and Gaelic}
\label{tab:atu-stats}
\end{table}

The second classification task was to predict the storyteller's gender. The gender of the narrator was strongly skewed towards male in both the Irish data (83.6\% male) and the Gaelic data (83.2\% male). The gender was reported as ‘unknown’ for 58 of the tales in the dataset and these were omitted from the classifier. 

\begin{table}[h!]
\centering
\begin{tabular}{lll}
\hline
                         & \textbf{Irish} & \textbf{Gaelic} \\ \hline
\textbf{Count} & 1814           & 2566            \\ \hline
\textbf{Male \%}         & 83.6           & 83.2            \\ \hline
\textbf{Female \%}       & 16.4           & 16.8            \\ \hline
\textbf{\begin{tabular}[c]{@{}l@{}}Length male\end{tabular}} &
  \begin{tabular}[c]{@{}l@{}}$\mu$=1887 \\ ($\sigma$ =1825)\end{tabular} &
  \begin{tabular}[c]{@{}l@{}}$\mu$=1200 \\ ($\sigma$ = 2603)\end{tabular} \\ \hline
\textbf{\begin{tabular}[c]{@{}l@{}}Length female \\ \end{tabular}} &
  \begin{tabular}[c]{@{}l@{}}$\mu$=1709 \\ ($\sigma$ = 1500)\end{tabular} &
  \begin{tabular}[c]{@{}l@{}}$\mu$=689 \\ ($\sigma$ = 905)\end{tabular} \\ \hline
\end{tabular}
\caption{Tales by Gender and Length}
\label{tab:gender-stats}
\end{table}

\subsection{Metrics}
The F1 score is a commonly used metric for classification tasks, based on precision and recall. However, in light of the imbalanced distribution of labels in both tasks, we opted to report weighted F1 score as the metric (Eq. \ref{eq:weighted-f1})
\begin{equation}
\label{eq:weighted-f1}
\text{Weighted F1} = \frac{\sum_{i=1}^{k} w_i \cdot \text{F1}_i}{\sum_{i=1}^{k} n_i}
\end{equation}

where \( w_i \) is the weight given by the number of true instances in class \( c_i \), and \( \text{F1}_i \) is the F1 score for class \( c_i \).

\subsection{Baselines}

For both tasks, we divided the data into 70\% for the training dataset and 15\% each for the validation and test set, ensuring that the distribution of languages in the splits was the same as the overall distribution in the dataset. 

We set up a dummy baseline for both tasks, which was to select the most frequent label in the training dataset (\textit{Magic} for the ATU task and \textit{Male} for the gender task) and predict this as the correct label for every item in the evaluation. We also built a support vector machine (SVM) in sci-kit learn \cite{scikit-learn}, with a simple feature set as input: a count of the tokens which appear in each text, and the term-frequency inverse document frequency (TF-IDF), another count of the tokens with a weighting scheme which gives a higher weight to tokens with lower frequency.

\subsection{Transformer Models}
We selected three multilingual models that include Irish and/or Gaelic in their training data: mBERT -- a version of the BERT model \cite{devlin2018bert} trained on Wikipedia in 104 languages, XLM-RoBERTa \cite{conneau2019unsupervised} which is trained on the Common Crawl in 100 languages and the Language-agnostic BERT Sentence Embedding model (LaBSE), which is trained on 109 languages, with 17 billion monolingual sentences and 6 billion bilingual sentence pairs \cite{feng2020language}. We also included an Irish monolingual model, gaBERT \citep{barry2021gabert}, which is trained on 171 million tokens of Irish data. To date, no monolingual LLM exists for Gaelic data, so we omit this comparison.

We finetuned these models by using a classification head on top of each model with a loss function of binary cross-entropy for the gender task and cross-entropy for the ATU task. All classifiers were trained for 3 epochs with a batch size of 16 and a learning rate of 2e-5. 

\subsection{Adaptations}
\subsubsection{Length}
A known limitation of many Transformer models is that they rely on a self-attention mechanism, whose time- and space-complexity scales quadratically with respect to the sequence length \cite{keles2023computational}. For this reason, most use a maximum sequence length of 512 tokens. An alternative to this is the Local, Sparse and Global (LSG) attention introduced by \citet{condevaux2023lsg}, which approximates self-attention for sequences up to 4096 tokens. As more than 1,500 of the tales in our corpus exceed the 512 token limit, we adapted all of the models mentioned in the previous section for comparison with their base forms. 

\subsubsection{Domain}
As both of the languages in our dataset are low-resource, and from the specific genre of the folktale, we continued finetuning the models on the Schools and Maclean data described in Section \ref{domain-data}. This is referred to as domain-adaptive pre-training (DAPT), and it occurs before adding the classification head. Each model was finetuned for 3 epochs, with a batch size of 32 and a learning rate of 2e-5.

\begin{table}[h!]
\centering
\begin{adjustbox}{max width=\columnwidth }
\begin{tabular}{lcccc}
\toprule
\textbf{Model} & \textbf{LSG} & \textbf{DAPT} & \textbf{F1 ATU} & \textbf{F1 Gender} \\ 
\midrule
gaBERT        & $\checkmark$      & $\checkmark$      & \textbf{0.69}          & \textbf{0.90}         \\ 
SVM          & -      & -          & 0.68         & \textbf{0.90}         \\ 
gaBERT          & $\checkmark$          & $\times$      & 0.67          & 0.89         \\ 
LaBSE         & $\checkmark$      & $\times$          & 0.65          & \textbf{0.90}        \\ 
LaBSE          & $\times$      & $\times$      & 0.62          & 0.89         \\ 
gaBERT          & $\times$          & $\checkmark$          & 0.58          & 0.88         \\ 
gaBERT         & $\times$          & $\times$      & 0.56          & 0.89         \\ 
mBERT          & $\checkmark$      & $\checkmark$      & 0.56          & 0.87         \\ 
LaBSE         & $\times$          & $\checkmark$          & 0.54          & 0.88         \\ 
LaBSE        & $\checkmark$          & $\checkmark$          & 0.53          & 0.84         \\ 
mBERT        & $\times$      & $\checkmark$      & 0.52          & 0.88         \\ 
RoBERTa          & $\checkmark$      & $\checkmark$          & 0.49          & 0.82         \\ 
RoBERTa        & $\checkmark$      & $\times$          & 0.48          & 0.80         \\ 
RoBERTa          & $\times$          & $\checkmark$      & 0.43          & 0.66         \\ 
RoBERTa        & $\times$          & $\times$      & 0.43          & 0.66         \\ 
mBERT          & $\checkmark$          & $\times$          & 0.43          & 0.88         \\ 
mBERT          & $\times$          & $\times$          & 0.40          & 0.81         \\ 
Dummy        & -         & -         & 0.37          & 0.73         \\ 
\bottomrule
\end{tabular}
\end{adjustbox}
\caption{Gender and ATU Classification Results in order of highest F1 on ATU task. Base transformer models are indicated with an $\times$  for LSG and DAPT}
\label{tab:f1_atu_sorted}
\end{table}

In terms of our first research question: the base models give very varied performance on this task. LaBSE performs best for Irish and Gaelic, followed by gaBERT, while mBERT and RoBERTa do not beat the dummy baseline for the gender task, and on the ATU prediction task, they failed to make any correct predictions of the two most infrequent labels - ‘Ordinary’ and ‘Animal’. 

Increasing the maximum sequence length improved the performance of the LaBSE and gaBERT models more than domain adaptation did. The combination of LSG and DAPT gave better results than either augmentation on its own for gaBERT, mBERT and RoBERTa, however this combination disimproved the performance of LaBSE over the base model. 

Arguably the most interesting result is the strong performance of the SVM. Although the input features were a bag of words and TF-IDF representation, they outperformed almost every model, except the length and domain-adapted gaBERT.

\section{Discussion}
In order to explore which multilingual LLMs give a faithful representation of two low-resource languages - Irish and Gaelic - we set up two classification tasks on a small folklore dataset, and measured the models’ performance under four conditions: base model, domain-adapted, length-adapted and both adaptations. Of the base models, the language-agnostic LaBSE and the Irish-only gaBERT models performed best. RoBERTa and mBERT did not beat a very simplistic baseline on one of the tasks, and performed poorly at the other. Surprisingly, one of the best results came from an SVM model with non-contextual features.  Of the two adaptations we presented, length augmentation improved our results more than domain adaptation, except in the case of LaBSE. The combination of both adaptations gave the best results for gaBERT, mBERT and RoBERTa.

Several interesting points arise from our experiments. Firstly, RoBERTa and mBERT are two of the most well-known transformer models, and their performance on our dataset indicates that they may not represent low-resource languages as well as well as they claim. Even with adaptations for domain and length, they achieved minimal improvements on the classification tasks. 

Secondly, it was not anticipated that gaBERT would perform so well, particularly in light of the fact that it is not trained on any Gaelic data. However, the model's input data was tokenized with a SentencePiece tokenizer, and it is possible that there was positive transfer between Irish and Gaelic due to a high number of shared sub-word tokens between the two Celtic languages. It is also notable that domain adaptation did not improve the model’s results much, it was the length augmentation that improve performance. However, as Table \ref{tab:gender-stats} indicates, tales told by men make up 83\% of the dataset, and with an average length of 1,200 - 1,887 words, it is plausible that allowing the model to process inputs up to 4,096 tokens was beneficial. 

Finally, the competitive performance of the SVM is an important takeaway from this paper. Although transformers have yielded state of the art results in NLP for some years now, the have a number of limitations, such as interpretability and energy consumption. On the latter point, the SVM classifier model trained in seconds, compared to ~4 minutes for a base or adapted LLM and 30+ minutes for an LSG model. Researchers should consider an SVM as their first choice of model due to its performance, ease of implementation and lower energy footprint. 

In order to further understand the performance of our best performing models, we examined the training curves of the SVM and gaBERT models to check for overfitting. We trained the models on subsets of 10\%, 30\%, 50\%, 70\% and 100\% of the data. After training on each subset, the model is evaluated on the same training data, in order to determine how well it performs on seen data. The validation scores measure the model's performance on unseen data. 

\begin{figure}[h]
\includegraphics[width=8cm]{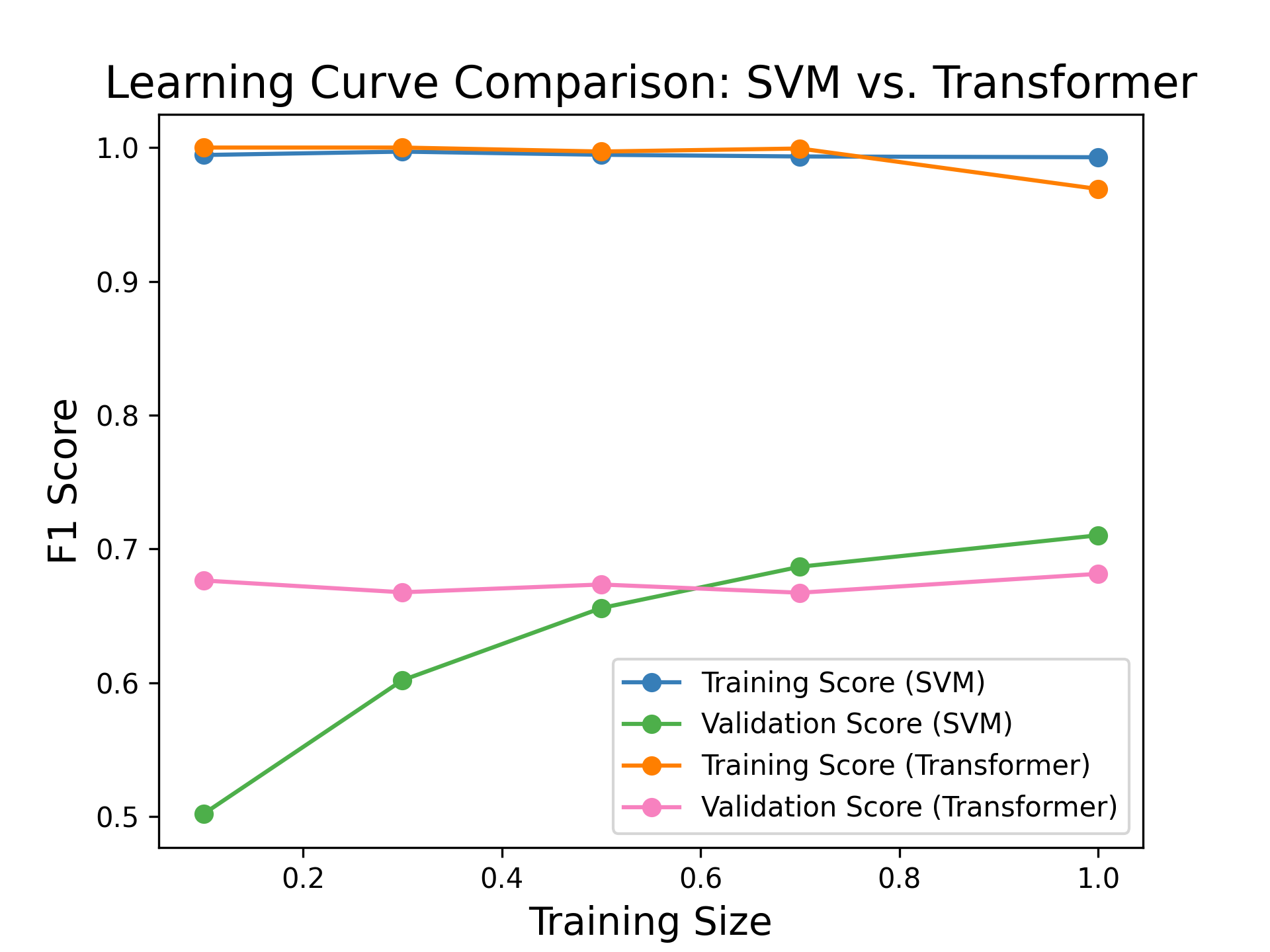}
\centering
\caption{Training Curves and Validation Scores for SVM and gaBERT on the ATU bin Prediction Task}
\label{fig:training-curve}
\end{figure}

Figure \ref{fig:training-curve} indicates that both the SVM and gaBERT are overfitting to the training data for ATU prediction, as we can see early on that the training score is high. The disparity between this score and the cross-validation scores for both models also shows that this overfitting means that the models do not generalize as well as hoped to unseen data. Furthermore, the transformer model performs almost as well on 10\% of the data as it does on the full dataset. 

\begin{figure}[h]
\includegraphics[width=8cm]{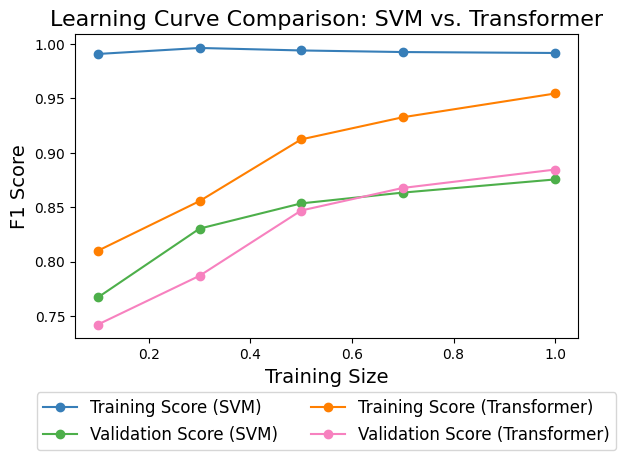}
\centering
\caption{Training Curves and Validation Scores for SVM and gaBERT on the Narrator Gender Prediction Task}
\label{fig:training-curve2}
\end{figure}

Similarly, in the gender task, the SVM overfits the training data, as does the transformer to a lesser extent. However, the transformer's validation score indicates that it generalises better to unseen data. The performance of the transformer seems to plateau with about 50\% of the data, while the SVM could possibly improve performance with some additional data. The use of regularization techniques, or a different kernel may lessen the extent of the overfitting.

\section{Conclusion}
We present work examining the performance on four LLMs on a classification task, along with three augmentations to the models. The base models do not perform as well as the augmented ones, with mBERT and RoBERTa failing to beat a dummy baseline. The best performing model was trained on Irish data only, was domain-adapted with continued finetuning and used LSG attention to increase the size of its context window. However, our linear baseline - an SVM with bag of words and TF-IDF features gave the second best performance. We conclude that the low-resource setting is challenging for LLMs, and while augmentations do help, classical machine learning models are still a competitive choice. 

\section*{Limitations}
The size of the dataset, at just over 4,600 items, is a limitation. However, it is somewhat characteristic of the low-resource setting. The skewness of the data is a second limitation, there were two labels in the ATU task and one in the gender task that were vastly under-represented, and although we tried to mitigate this through the use of the weighted F1 score, this coupled with the size of the dataset is challenging for large models. Finally, we relied exclusively on extrinsic evaluation, i.e. classification performance. Future work will look at intrinsic measure of LLM's ability to represent a language, e.g. pseudo-log likelihood.

\section*{Ethics Statement}
Institutional ethical review for this research was granted by the Ethics Officer of the School of Literatures, Languages and Cultures, University of Edinburgh. No substantial risks are associated with it.


\bibliography{custom}
\bibliographystyle{acl_natbib}



\end{document}